\documentclass{neus2025}

\usepackage{enumitem}

\usepackage{booktabs}

\usepackage{xy} \input xy \xyoption{all}

\usepackage{tikz}
\usepackage{tikz,ifthen}
\usetikzlibrary{shapes.geometric, arrows}

\usepackage{listings}
\usepackage{xcolor}

\usepackage{enumerate}

\usepackage{algorithm}
\usepackage{algorithmic}
 \usepackage{float} 

\newtheorem{thm}{Theorem}[section]
\newtheorem{lem}{Lemma}

\usepackage[capitalise]{cleveref}
\crefname{thm}{Thm.}{}
\crefname{prop}{Prop.}{}
\crefname{lem}{Lem.}{}
\crefname{cor}{Cor.}{}
\crefname{prob}{Problem}{}
\crefname{figure}{Fig.}{}
\crefname{exa}{Example}{}
 \crefname{table}{Table}{}

\newcommand\p{\mathfrak p}
\newcommand\F{\mathbb F}
\newcommand\Z{\mathbb Z}
\newcommand\Q{\mathbb Q}

\DeclareMathOperator\Gal{Gal }	
\DeclareMathOperator\Stab{Stab }	
\DeclareMathOperator\h{H}					

\newcommand\G{\mathcal G}
\newcommand\cP{\mathcal P}

\title[Neuro-Symbolic Learning for Galois Groups]{Neuro-Symbolic Learning for Galois Groups: Unveiling Probabilistic Trends in Polynomials}
\usepackage{times}

\author{%
 \Name{Elira Shaska} \Email{elirashaska@oakland.edu}\\
 \addr Department of Computer Science, Oakland University, Rochester, MI, 48309.
 \AND
 \Name{Tony Shaska} \Email{shaska@oakland.edu}\\
 \addr Department of Mathematics and Statistics, Oakland University, Rochester, MI, 48309.
 }

\begin{document}

\maketitle

\begin{abstract}
This paper presents a neuro-symbolic approach to classifying Galois groups of polynomials, integrating classical Galois theory with machine learning to address challenges in algebraic computation. By combining neural networks with symbolic reasoning—leveraging invariants like root distributions, signatures, and resolvents—we develop a model that outperforms purely numerical methods in accuracy and interpretability. Focusing on sextic polynomials with height \(\leq 6\), we analyze a database of 53,972 irreducible examples, uncovering novel distributional trends, such as the 20 sextic polynomials with Galois group \(C_6\) spanning just seven invariant-defined equivalence classes. These findings offer the first empirical insights into Galois group probabilities under height constraints and lay the groundwork for exploring solvability by radicals. Demonstrating AI’s potential to reveal patterns beyond traditional symbolic techniques, this work paves the way for future research in computational algebra, with implications for probabilistic conjectures and higher-degree classifications.
\end{abstract}


\section{Introduction}

Galois theory, a cornerstone of modern algebra, elucidates whether polynomial equations can be solved using radicals, offering profound insights into the symmetry of their roots. Since Évariste Galois’s seminal work in the 19th century, it has been established that polynomials of degree five or higher are generally unsolvable by radicals—a result rooted in the properties of their Galois groups. However, if a polynomial’s Galois group is solvable, radical solutions exist, expressible through formulas derived from the group’s solvable towers, a sequence of subgroups leading to a solvable extension. For lower-degree polynomials, classical tools like discriminants and resolvents often suffice to identify these groups, yet for higher degrees, the escalating complexity demands more intricate computational methods. Despite extensive literature on computational Galois theory, the field continues to grapple with challenges, such as efficient group classification and understanding their distribution across polynomial families.

Could artificial intelligence (AI) and machine learning provide fresh perspectives on these enduring problems? This paper seeks to lay the foundation for addressing this question, exploring how AI might enhance our ability to probe Galois theory’s depths. What tasks could such an AI model perform, and how effectively could it tackle these challenges? To formalize this inquiry, let \(\cP_n\) denote the set of all degree \(n\) monic irreducible polynomials in \(\mathbb{Z}[x]\). For a polynomial \(f \in \cP_n\), expressed as \(f(x) = \sum_{i=0}^n a_i x^i\), we define its height as \(\h(f) = \max \{ |a_i| \}\), its Galois group over \(\mathbb{Q}\) as \(G = \Gal_{\mathbb{Q}}(f)\), and \(\G_n\) as the list of transitive subgroups of \(S_n\). An ideal model would achieve the following:
\begin{itemize}[noitemsep, topsep=0pt]
\item Given \(f \in \cP_n\), determine \(G \in \G_n\) such that \(G = \Gal_{\mathbb{Q}}(f)\).
\item For any \(G \in \G_n\), find the polynomial \(f\) with minimal height such that \(\Gal_{\mathbb{Q}}(f) = G\).
\item For \(G \in \G_n\), estimate the probability that a polynomial \(f \in \cP_n\) with height \(\leq H\) has \(\Gal_{\mathbb{Q}}(f) = G\).
\item For \(f \in \cP_n\) with a solvable \(\Gal_{\mathbb{Q}}(f)\), derive radical formulas to express its roots.
\item For each \(G \in \G_n\), compute resolvents to aid in group identification.
\end{itemize}

In this work, we introduce a neuro-symbolic approach to classifying Galois groups, integrating machine learning with algebraic techniques. We leverage neural networks to identify patterns while embedding symbolic reasoning to ensure mathematical accuracy. By constructing datasets of polynomials with known Galois groups and incorporating symbolic layers—such as those based on root counts, signatures, and resolvents—we demonstrate how AI can augment algebraic computations, enhancing both precision and interpretability over purely numerical methods.

We focus on polynomials of degree six (sextics) because they strike a balance between complexity and tractability, making them an ideal testbed for our approach. Sextic polynomials are particularly appealing due to their well-documented algebraic invariants and the availability of extensive databases of binary sextics, such as those derived from genus two curves \cite{2024-03}. These resources enable us to rigorously test and refine our model by comparing its predictions against established classifications, providing a concrete foundation for evaluating its effectiveness.

A persistent challenge in applying machine learning to mathematical domains like Galois theory is ensuring adherence to rigorous principles. Standard machine learning methods often falter due to the scarcity of large, labeled datasets and the intricate nature of algebraic structures. Without symbolic reasoning, these models risk missing critical constraints—such as the link between a polynomial’s discriminant and its Galois group’s parity—or failing to generalize from sparse data, leading to predictions that violate theorems or lack meaning. Neural networks excel at pattern recognition but struggle to capture the deep logical consistency required, a gap exacerbated by the computational difficulty of generating labeled Galois group data.

To overcome these limitations, we employ a neuro-symbolic framework that fuses machine learning with algebraic reasoning. Our model incorporates mathematical invariants, such as root distributions and modular reduction properties, to guide the learning process, enabling it to draw from both empirical data and theoretical principles. This hybrid approach mitigates erroneous patterns, focusing on algebraically significant structures, and proves more reliable for polynomial classification, as evidenced by our sextic case study.

Beyond classification, our analysis of 53,972 sextic polynomials with height \(\leq 6\) unveils novel probabilistic insights, quantifying the occurrence of Galois groups for the first time under such constraints. This empirical dataset reveals striking trends—for instance, the cyclic group \(C_6\) corresponds to just five \(\mathrm{GL}_2(\mathbb{Q})\)-equivalence classes among 20 polynomials—suggesting a structured sparsity that challenges existing expectations. These findings not only validate our model but also offer a fresh lens on the distribution of Galois groups, a question with deep ties to probabilistic Galois theory.

By merging Galois theory’s classical elegance with AI’s modern capabilities, this work pioneers a path for solving longstanding mathematical problems. Our neuro-symbolic approach, tested on sextics, demonstrates enhanced classification accuracy and uncovers distributional patterns inaccessible to traditional methods alone. While challenges like scaling to higher degrees remain, these results—grounded in a robust framework and original data—open new horizons for computational algebra and AI-driven mathematical discovery, potentially informing conjectures about group frequencies and beyond.

\section{Galois groups of  polynomials}\label{sect-2}
Let $\F$ be a perfect field.  For simplicity we only consider the case when ${\mathrm{char }} \F=0$.  
Let $ f(x) $ be a degree $n=\deg f$  irreducible polynomial in $ \F[x]$ which is factored as $f(x)=  \prod_{i=1}^n (x-\alpha_i)$
in a splitting field $E_f$. Then, $E_f/\F$ is Galois because it is a normal and separable extension. The group  $ \Gal (E_f/\F)$ is called \textbf{the Galois group} of $ f(x)$ over $\F$ and  denoted by $\Gal_\F (f)$. The elements of $\Gal_\F (f)$  permute roots of $ f(x)$. Thus, the Galois group of polynomial has an isomorphic  copy embedded  in $S_n$, determined up to conjugacy by $f$.
If  $f(x)\in \F[x]$ is  irreducible,  then  $\deg f \, \mid  \, |G|$. 
Denote by $\Delta_f$ the discriminant of $f$ and   $G=\Gal_\F (f)$. Then:

i)   $G\cap A_n= \Gal (E_f / \F(\sqrt{\Delta_f}))$. In particular, $G \subset A_n$ if and only if    $\sqrt{\Delta_f} \in \F$.

ii) The irreducible factors of $f$ in $\F[x]$ correspond to the orbits of $G$. In particular, $G$ is a transitive subgroup of $S_n$ if and only if  $f$ is irreducible.
 
 Indeed,  the Galois group  $\Gal_\F (f)$ is an affine invariant of $f(x)$  (e.g.,  invariant under affine transformations).  In other words, $\Gal (f)\cong\Gal(g)$ for any  $g(x)=f(ax+b)$,     for $a, b \in \F$ and $a\neq 0$.
 
Let $f(x) \in {\mathbb Q}[x]$ be given by  $f(x)=\sum_{i=0}^n a_i x^i$. 
Let $d$ be the common denominator   of all coefficients $a_0, \ldots, a_{n-1}$. Then $g(x):=d \cdot f(\frac x d)$ is a monic polynomial with integer coefficients. Clearly the splitting field of $f(x) $ is the same as the splitting field of $g(x)$. Thus, without loss of generality we can assume that $f(x) \in {\mathbb Z}[x]$ is a monic polynomial
with integer coefficients.

Using computational group theory and GAP, we can compute a list of transitive subgroups for relatively large $n$.  
These precompiled lists for every $n$ will be our candidates for Galois groups.  

\subsection{Reduction modulo a prime and signature of groups}
Factoring $f(x)$ modulo a prime $p$ can give a lot of information about the Galois group due to the following theorem of Dedekind; see   \cite[section 8.10]{warden} for a proof. 

\begin{thm}\textbf{(Dedekind)} 
Let $f(x) \in {\mathbb Z}[x]$ be a monic polynomial such that  $\deg f = n$, $\Gal_{\mathbb Q} (f) = G$, and $p$ a prime such that $p \nmid {\Delta}_f$. If $f_p:=f(x) \mod p \, \, $ factors in ${\mathbb F}_p [x]$ as a product of irreducible factors of degree     $n_1, n_2, n_3, \cdots, n_k$, 
then $G$ contains a permutation of type     $(n_1)\, (n_2) \, \cdots \, (n_k)$
\end{thm}

Hence, we can determine the Galois group in many cases since the \emph{type} of permutation in $S_n$ determines the conjugacy class in $S_n$.   
The \emph{signature} of $G$ is the set of such types of permutations.

\subsection{Polynomials with non-real roots}\label{real-roots}
Denote by $r$ the number of non-real roots of $f(x)$. Since the complex conjugation permutes the roots,  then $r$ is even, say $r=2s$. 
By reordering the roots, we may assume that if  if $f(x)$ has $r$ non-real roots then   $\alpha   :=(1, 2) (3, 4)\cdots (r-1, r) \in \Gal (f)$.    
The existence of $\alpha$ can narrow down the list of candidates for $\Gal (f)$. However, it is unlikely that the group can be determined 
based solely on this information
 unless $p$ is "sufficiently large.  

\begin{thm}[ \cite{2004-1}]
Let $f(x)\in {\mathbb Q}[x] $ be an irreducible polynomial of prime degree $p \geq 5$ and  $r=2s$ be the
number of non-real roots of $f(x)$. If $s$ satisfies     $s \, ( s \log s + 2 \log s + 3) \leq p$,     then $\Gal (f)$ is isomorphic to either  $A_p$ or  $S_p$.
\end{thm}

Denote the above bound on $p$ by   $N(r) :=   s \, ( s \log s + 2 \log s + 3)$,  for $r=2s$. Hence, for a fixed number of non-real roots, for  $p \geq N(r)$ the Galois group is always $A_p$ or $S_p$.
Hence, for  a polynomial of prime degree $p$ which has $r$ non-real roots, the Galois group  $\Gal (f) = A_p$ or $S_p$ if one of the following  holds:
i)  $r=4$ and $p > 7$; ii)  $r=6$ and $p > 13$; iii)  $r=8$ and $p > 23$; iv)  $r=10$ and $p > 37$. 
These    results have been generalized to every degree (not necessary prime) , but the result is more technical to be stated here. 

\subsection{A word on equivalence classes of polynomials}
Following classical invariant theory, we represent polynomials in their binary form.  
To every polynomial $f(x)$ we associate a binary form  $f(x, y)=y^n f\left(  \frac x y \right)$
 which  is called the \textit{homogenization of $f(x)$}.  Conversely, every binary form $f(x, y)$  can be associated to a polynomial $f(x, 1)$, called the      \textit{dehomogenization of $f(x, y)$}.  

Two degree $n$ binary forms $f, g \in \Z [x, y]$ are called  ${\mathrm{GL}}_2 ( \Z)$-equivariant   if $g(x, y) = \pm f(ax+by, cx+dy)$ for some 
$\begin{bmatrix}
a & b \\ c & d 
\end{bmatrix}  \in {\mathrm{GL}}_2 (\Z)
$.   
Two degree $n$ polynomials $f, g \in \Z[x]$ are called \textbf{${\mathrm{GL}}_2 (\Z)$-equivalent} if their homogenizations are ${\mathrm{GL}}_2 (\Z)$-equivalent.

\subsection{Resolvents}
For a basic reference see \cite[Sec. 6.3, p. 322 ]{cohen}.  Let $G\leq S_n$ such that it contains $\Gal (f)$. For $F\in \Z[x_1, \ldots , x_n]$   the stabilizer of $F$ in $G$ is defined as
\[
\Stab_G (F) \, = \, \{   
\sigma \in G \, \mid \, F( x_{\sigma (1)}, \ldots , x_{\sigma(n)} ) = F(x_1, \ldots , x_n)
\}
\]
We denote $H:=\Stab_G (F)$ and by $G/H$ and set of left coset representatives of $H$ in $G$.  
The \textbf{resolvent polynomial} $R_G (F, f)$    with respect to $G$, $F$, and $f(x)$ is defined as 
\[
R_G (F, f) (x)  \, = \, \prod_{\sigma \in G/H} \left(  x-F  ( \alpha_{\sigma(1)},  \ldots ,  \alpha_{\sigma(n)}    )    \right)
\]
Then $R_G (F, f) (x)\in \Z[x]$  and $m:=\deg R_G (F, f) (x) = [G:H]$. Then there is a group homomorphism $\phi : G \to S_m$ given by the natural left action of $G$ on $G/H$. 

\begin{thm}
If $R_G (F, f) (x)$ is squarefree then $\Gal \left(  R_G (F, f) (x)\right) \cong \phi \left( \Gal (f) \right)$. 
Moreover, the list of degrees of irreducible factors of  $R_G (F, f) (x)$ is the same as  the list of lengths of the orbits on the action of $\phi (\Gal (f))$ on $[1, \ldots , m]$. 
\end{thm}

Notice that if $R_G (F, f) (x)$ is note squarefree we can use apply a Tschirnhausen transformation to transform the polynomial to a lower degree polynomial; see \cite[Algorithm 6.3.4]{cohen}. 

When we compute a group, we will output not only the isomorphism class
of the group, but also a sign expressing whether the group is contained in $A_n$
(+ sign) or not (- sign). This will help resolve a number of ambiguities since
isomorphic groups are not always conjugate in $S_n$. 

Note that one resolvent is enough to determine the Galois group for quartics and quintics, while at least two resolvents are required to do so for sextics. 

\subsection{Probabilistic Galois Theory}\label{prob-galois}
 Let $E_n (H)$ denote the cardinality of the set of monic integer polynomials of degree $n$ having height at most $H$, whose Galois group is not the full symmetric group $S_n$:
\[
    E_n (H) := \#\{ f  \in \Z[x]  : \h (f)  \leq H   \text{ and }    \Gal_{\Q} (f)    \text{ is not  } S_n\}.
\]
In 1936, B. L. van der Waerden proved that
\[
    E_n(H) \ll_n H^{n - 1 / 6(n-2) \log\log H}.
\]
Its proof relied on considering reductions modulo prime numbers and a sieve argument. There are many papers on this topic and the above bound has been improved by many authors. 
In \cite{dietmann} the author uses Galois resolvents proves that for $\varepsilon > 0$, 
\[
  E_n(H) \ll_{n, \varepsilon}  H^{n - 2+ \sqrt{2} + \varepsilon}.
\]
Furthermore, if     $E_n^G (H) $ is  the number of polynomials with height $\leq H$ and Galois group $G$, and $\delta_G = 1/|S_n/G|$, then 
\[
\# E_n^G (H)  \ll_{n, \varepsilon} H^{n-1+\delta_G+\varepsilon}.
\]

\section{Irreducible sextics: a case study}
We now apply the methods outlined above to the case of sextic polynomials, which serve as an ideal test case for our neuro-symbolic approach. Sextics strike a balance between algebraic complexity and data availability, making them suitable for both theoretical analysis and empirical validation. Their Galois groups, while complex enough to challenge purely numerical methods, are well-studied and supported by extensive databases of binary sextics, such as those derived from the study of genus two curves \cite{2024-03}. Moreover, resolvent-based techniques, as detailed in \cite{king, solv-sextics}, provide a benchmark for comparison, allowing us to illustrate how a neural network framework can enhance or complement traditional algebraic methods in determining Galois groups.

A binary sextic is given by   $f(x, y)  = \sum_{i=0}^6 a_i x^i y^{6-i} \in \Q[x, y]$.  As noted above $[f]=[a_0: \ldots : a_6]\in {\mathbb P}_\Q^6$. The height of $f$ is the projective height of $[f] \in {\mathbb P}_\Q^6$.   We generate datasets of irreducible sextic polynomials with bounded heights using 
  \textsl{Sagemath}:
\begin{verbatim}
PP = ProjectiveSpace(6, QQ)
rational_points = PP.rational_points(h)
\end{verbatim}

There are   3 280 326  rational points with height $\leq 6$ in ${\mathbb P}^6$ of which 53,972 represent irreducible polynomials.  
In the second and third columns of \cref{table-2} are given all proper transitive subgroups of $S_6$ and  their Gap identities.  For the purposes of this paper and our database, we will label them as $g1, \ldots , g15$ as in the first column of \cref{table-2}. 

\begin{table}[h!]
    \centering
    \begin{tabular}{@{}ccccccc@{}}
        \toprule
        \# & \textbf{GroupId} & \textbf{Group}  	& $\mathrm{sig}$	&	$E_6(4)$ 	&	$E_6(5)$ 	&  $E_6(6)$  \\ 
        \midrule
        1  & [6, 2]          & \( C(6)   \) 		&	[0,0,1,0,0,1,0,0,0,1] 		&	12	&	&  20  \\
        2  & [6, 1]          & \( D_6(6)   \) 		& 	[0,0,1,0,0,1,0,0,0,0]		&	25	&	&  43  \\ 
        3  & [12, 4]         & \( D(6)   \)  		&  	[0,0,0,0,0,0,0,0,0,0]		&	402	&	&  1 185  \\ 
        4  & [12, 3]         & \( A_4(6)   \) 		&	[0,0,0,0,0,0,0,0,0,0]		&	18	&	&  34  \\   
        5  & [18, 3]         & \( F_{18}(6)   \) 	&  	[0,0,1,1,0,1,0,0,0,1]		&	124	&	&  222  \\  
        6  & [24, 13]        & \( 2A_4(6)  \) 	& 	[1,1,1,0,0,1,0,0,0,1]		&	192	&	&  394  \\ 
        7  & [24, 12]        & \( S_4(6d)   \) 	&	[0,1,0,0,0,1,0,1,0,0]		&	581	&	&  2 608  \\
        8  & [24, 12]        & \( S_4(6c)   \) 	&	[0,1,1,0,0,1,1,0,0,0]		&	42	&	& 128   \\   
        9  & [36, 10]        & \( F_{18}(6):2   \) 	& 	[0,1,0,1,0,1,0,1,0,0]		&	170	&	& 648   \\ 
        10 & [36, 9]         & \( F_{36}(6)   \) 	& 	[0,1,1,1,0,1,0,0,0,1]		&	18	&	& 58   \\ 
        11 & [48, 48]        & \(   2 \wr S(3) \) 	&	[1,1,1,0,0,1,1,1,0,1]		&	4367	&	& 20 236   \\ 
        12 & [60, 5]         & \(  PSL(2,5)   \) 	& 	[0,1,0,0,0,1,0,0,1,0]		&	264	&	&  706  \\ 
        13 & [72, 40]        & \( S(3) \wr 2 \) 	&  	[1,1,1,1,1,1,0,1,0,1]		&	7616&	&  26 024  \\ 
        14 & [120, 34]       & \(  PGL(2,5)   \) 	& 	[0,1,1,0,0,1,1,0,1,1]		&	160	&	&  534  \\ 
        15 & [360, 118]      & \( A_6 \) 		&	[0,1,0,1,0,1,0,1,1,0]		&	264	&	&  1 092  \\ 
        \bottomrule
    \end{tabular}
    \caption{Transitive Subgroups of \( S_6 \) and their occurrences in the database for each height}
    \label{table-2}
\end{table}

\begin{figure}[hb!]  
   \centering
   \includegraphics[width=4in]{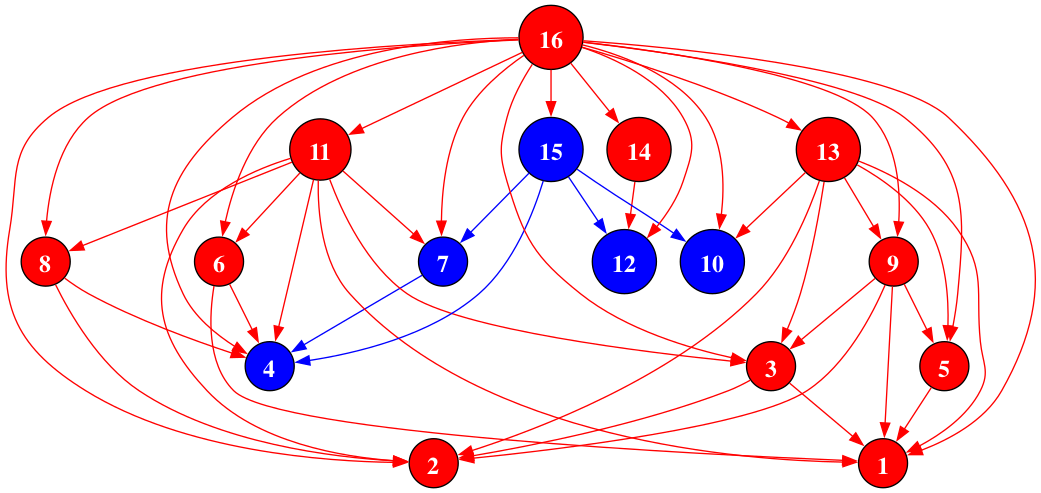} 
   \caption{Lattice of transitive subgroups of $S_6$ with blue nodes representing subgroups of $A_6$.}
   \label{fig:s6}
\end{figure}

%
We label the above conjugacy classes above (not counting identity) as
$C_1=(2)$,  $C_2=(2)^2$,  $C_3=(2)^3$,
 $C_4=(3)$,
  $C_5=(3)(2)$,
   $C_6=(3)^2$,
    $C_7=(4)$,
     $C_8=(4)(2)$,
      $C_9=(5)$,  $C_{10}=(6)$. 
Define the \textbf{signature} of $G$ as the 10-tuple 
$
\mbox{sig} (G) = [  \alpha_1, \ldots , \alpha_{10} ]
$,
where $\alpha_i=1$ if $G$ has elements in $C_i$ and $\alpha_i=0$ otherwise.    
In \cite[Algorithm 6.3.10]{cohen} it is describes how the resolvent method can be used to determine the Galois group of sextic polynomials. 
These subgroups can be visualized in the  lattice diagram \cref{fig:s6} which organizes them based on containment. Groups with blue color corresponds to the subgroups of $A_6$. 
Notice that the digram in \cref{fig:s6} also provides and outline of the algorithm that can be used in determining the Galois group of sextics and this is true for any degree. 
  
%

\subsection{Counting polynomials}
From our brief description of probabilistic Galois theory in \cref{prob-galois} we have estimates on how often a group should occur for a given bound.   In the last columns of \cref{table-2} we have shown such occurrences for each height $h\leq 6$.  Next, and probably most interesting part of the data is the distribution of data among the groups.  

\begin{lem}
There are exactly 53 972 irreducible sextics with height $h \leq 6$  which  have Galois group not isomorphic to $S_6$.  The list of such groups and their frequencies are presented in \cref{table-2}.
\end{lem}

Moreover,   20 sextics $f(x)$ with $\Gal (f) \cong C_6$ and their ${\mathrm{SL}}_2 (\Q)$-invariants are as in \cref{tab:polynomials}. 
\begin{small}
\begin{table}[h!]
\begin{center}
\begin{tabular}{|c|c|c|}
\hline
$\#$  	& $f(x) $  								& $\p$ \\
\hline
1     &    $ x^6 - x^3 + 1 $      						&    $  [-234, 1944, -129762, -19683]    $     \\ 
2     &    $ x^6 + x^3 + 1 $      						&             \\ 
\hline
3     &    $ x^6 - x^5 + x^4 - x^3 + x^2 - x + 1 $      		&   $   [-210, 1176, -76146, -16807]     $    \\ 
4     &    $ x^6 + x^5 + x^4 + x^3 + x^2 + x + 1 $      		&              \\ 
\hline
5   &    $ x^6 + 2 x^5 + 5 x^4 + 3 x^2 + x + 1 $      		&              \\ 
6    &    $ x^6 - 2 x^5 + 5 x^4 + 3 x^2 - x + 1 $      		&   $   [-400, 6076, -315952, -10955763]     $    \\ 
7     &    $ x^6 + x^5 + 3 x^4 + 5 x^2 + 2 x + 1 $      		&               \\ 
8     &    $ x^6 - x^5 + 3 x^4 + 5 x^2 - 2 x + 1 $      		&             \\ 
\hline
9     &    $ x^6 + 2 x^5 + 4 x^4 + x^3 + 2 x^2 - 3 x + 1 $      &               \\ 
10     &    $ x^6 - 2 x^5 + 4 x^4 - x^3 + 2 x^2 + 3 x + 1 $      &    $  [-602, 14896, -2453136, -1075648]   $      \\ 
11     &    $ x^6 + 3 x^5 + 2 x^4 - x^3 + 4 x^2 - 2 x + 1 $      &                \\ 
12     &    $ x^6 - 3 x^5 + 2 x^4 + x^3 + 4 x^2 + 2 x + 1 $      &               \\ 
\hline
13 & 		$x^6 + 6 x^4 + 5 x^2 + 1 $   				 &   $ [-720,6468,-1435896, -153664]  $       \\ 
14 & 	$	x^6 + 5 x^4 + 6 x^2 + 1 $    &     \\ 
\hline
15	&	$-x^6 - x^5 + 5 x^4 + 4 x^3 - 6 x^2 - 3 x + 1 $    &     \\ 
16	&	$-x^6 + x^5 + 5 x^4 - 4 x^3 - 6 x^2 + 3 x + 1 $    &  $ [936,10140,2926404, 371293]   $ \\ 
17	&	$-x^6 - 3 x^5 + 6 x^4 + 4 x^3 - 5 x^2 - x + 1 $    &    \\ 
18	&	$-x^6 + 3 x^5 + 6 x^4 - 4 x^3 - 5 x^2 + x + 1 $    &      \\ 
\hline
19	&	$3 x^6 - 6 x^3 + 6 x^2 - 3 x + 1 $    			&  $ [-504,22356,-3327156, -26946027] $   \\ 
20	&	$3 x^6 + 6 x^3 + 6 x^2 + 3 x + 1 $    &       \\ 
\hline
\end{tabular}
\end{center}
    \caption{The only sextics with height $H \leq 6$ and Galois group   $C_6$}
\label{tab:polynomials}
\end{table}%
\end{small}
 
\subsection{Data quality versus data quantity}  

The main question is: Is it worth it to have a database with all irreducible polynomials or simply their invariants?  Our motivation comes from \cite{2023-01}, \cite{2024-03}

The ${\mathrm{GL}}_2 (\Q)$-equivalence  class of  $f(x, y)$ is determined by the invariants of $f(x, y)$, which are commonly denoted by $J_2(f)$, $J_4(f)$, $J_6(f)$, $J_{10} (f)$ and are homogenous polynomials of degree 2, 4, 6, and 10 respectively in the coefficients of $f( x, y)$.
Hence, the  moduli space of binary sextics   is isomorphic to the weighted projective space ${\mathbb P}_{(2,4,6,10)}$. 

There is however the possibility that two different points $\p$ corresponds to the same moduli class because such points are up to equivalence in $\mathbb{P}_{(2,4,6,10)} (\Q)$.  
Next we will explain how to count such points in the weighted projective space.  
The following  invariants defined by Igusa   in \cite{Ig},
$ t_1= \frac {J_2^5} {J_{10}}$,   $t_2 = \frac {J_4^5} {J_{10}^2}$,  $t_3 = \frac {J_6^5} {J_{10}^3}$
 are defined everywhere in the moduli space and are ${\mathrm{GL}}_2$-invariants.    
Two binary forms are $GL_2 (\Q)$-equivalent if and only if they have the same absolute invariants. 
By using absolute invariants $t_1$, $t_2$, $t_3$ we can count the points in the moduli space and there are 25, 853 such points.

\section{Neurosymbolic Networks for Galois Group Determination}
Our first experiment aimed to determine the Galois group of polynomials. Since the overwhelming majority of irreducible polynomials have a Galois group isomorphic to \(S_n\) (e.g., \(S_6\) for degree 6), we focused only on cases where the group deviates from \(S_n\). We experimented with various machine learning models, including artificial neural networks, K-means clustering, and random forest classifiers, to classify the Galois groups of irreducible polynomials in our database. We anticipated modest results, but were surprised that no model achieved accuracy above 10\%. This outcome confirmed a mathematically unsurprising observation: the coefficients of polynomials alone lack detectable patterns sufficient to determine the Galois group. The only promising alternative was to integrate symbolic reasoning with numerical learning, leading us to develop a neuro-symbolic neural network.

\subsection{Symbolic Layers: Encoding Mathematical Knowledge}
Mathematically, several techniques can determine the Galois group, as outlined in Section~\ref{sect-2}: i) the number of real roots, ii) the signature, iii) the discriminant, and iv) resolvents. While other methods exist, we focused on these four for this study, with plans to explore additional approaches in future work.

\noindent \textbf{Real Roots Layer:}
If a polynomial has sufficient real roots, Section~\ref{real-roots} indicates the group is either \(A_d\) or \(S_d\). Computing real roots is typically efficient using numerical methods, such as Sturm’s theorem, which constructs a Sturm sequence from the polynomial \(f(x)\) and its derivative. By evaluating sign changes at large finite bounds (e.g., \(\pm 10^{10}\)), this method approximates the count of real roots over the real line, making it effective for polynomials with integer or rational coefficients.

\noindent \textbf{Signature Layer:}
The first symbolic layer computes the signature by factoring the polynomial \(f(x)\) modulo small primes (e.g., \(p = 2, 3, 5, 7\)). For each input key \((a_0, \dots, a_d)\), the signature \(\sigma(key)\) is compared against known signatures for degree \(d\). The set of possible groups (\(grps\)) is updated to include only those admitting this signature. If \(|grps| = 1\), the Galois group is uniquely determined, concluding the process.

\noindent \textbf{Discriminant Layer:}
The discriminant \(D\) is precomputed for all polynomials but factored only when necessary. This layer activates if candidate groups might or might not be contained in \(A_d\), serving as a final refinement step due to its computational cost.

\noindent \textbf{Invariants and Resolvents:}
For lower degrees (up to 11), invariants of binary sextics are well-known, but higher degrees require further investigation. For specific cases (e.g., group \(g6\) in Table~\ref{table-2}), resolvents like the degree-30 polynomial \(R_G(F)(x)\) (as detailed in \cite{resolvent}) provide additional constraints, though we omit its explicit form here.

\subsection{The Galois Network: A Neurosymbolic Architecture}
We designed the GaloisNetwork to integrate numerical learning with symbolic reasoning, processing polynomial coefficients and leveraging mathematical insights to predict Galois group labels. This hybrid approach combines deep learning’s pattern recognition with domain-specific rules, ensuring both accuracy and interpretability.

\begin{algorithm}[H]
\caption{Neuro-Symbolic Galois Group Determination}
\label{alg:neuro-symbolic-galois}
\begin{algorithmic}[1]
\STATE \textbf{Input:} Raw polynomial coefficients
\STATE \textbf{Output:} Predicted Galois group from refined set \(grps\)

\STATE Initialize \(grps \leftarrow\) all transitive subgroups of \(S_n\)
\STATE \COMMENT{Symbolic Layer 1: Real Roots}
\IF {sufficient real roots are detected}
    \STATE Refine \(grps \leftarrow \{A_n, S_n\}\)
\ENDIF
\STATE \COMMENT{Symbolic Layer 2: Discriminant}
\STATE Compute discriminant \(D\)
\STATE Update \(grps\) based on \(D\)'s properties
\STATE \COMMENT{Numerical Layer 1}
\STATE Process coefficients and symbolic outputs using a dense neural network
\STATE \COMMENT{Symbolic Layer 3: Signature}
\STATE Determine polynomial's signature
\STATE Further constrain \(grps\) based on signature
\STATE \COMMENT{Numerical Layer 2}
\STATE Extract advanced features from combined symbolic and numeric data
\STATE \COMMENT{Fusion Layer}
\STATE Integrate all features into a singular representation
\STATE \COMMENT{Output Layer}
\STATE Predict Galois group classification over refined \(grps\)
\STATE \COMMENT{Post-Processing}
\STATE Apply algebraic constraints to ensure classification accuracy
\STATE \textbf{Return} Predicted Galois group
\end{algorithmic}
\end{algorithm}

The network takes feature vectors of polynomial coefficients as input. Additional features, such as root counts and Galois group characteristics, are computed using mathematical invariants, enriching the representation of each polynomial.

The GaloisNetwork is a fully connected feedforward neural network with an input layer matching the feature vector size, three hidden layers of 64 neurons each (using ReLU activation), and an output layer producing a probability distribution over possible Galois groups via softmax activation. This architecture captures complex relationships between features and group classifications.

Training minimizes cross-entropy loss using the Adam optimizer over 100 epochs, with periodic validation to monitor loss and refine learning. A post-processing step applies domain-specific rules, such as adjusting predictions based on real root counts, ensuring alignment with mathematical principles and enhancing reliability.

\textbf{Evaluation and Conclusion:} We assessed the GaloisNetwork using accuracy metrics, confusion matrices, and classification reports, demonstrating that integrating numerical and symbolic methods significantly improves performance over purely numerical approaches. This hybrid model not only achieves higher accuracy but also respects the mathematical structure of Galois theory, offering a robust and interpretable tool for polynomial analysis. Future work will explore additional invariants and higher-degree cases to further enhance its capabilities.

\section{Conclusions}

This paper demonstrates that a neuro-symbolic approach effectively classifies Galois groups of polynomials, surpassing purely numerical methods by integrating algebraic invariants with machine learning. Our case study on sextic polynomials uncovers distributional trends consistent with probabilistic Galois theory, highlighting the potential of artificial intelligence (AI) to address classical mathematical challenges. While obstacles persist—such as scaling to higher degrees and refining symbolic layers—this work lays a foundation for future research in computational algebra and AI-driven mathematical discovery. Future efforts will focus on exploring additional invariants, expanding datasets, and validating the model across broader polynomial classes.

To promote transparency and foster further investigation, we will make the detailed results, datasets, and Python code from this work publicly available. This release will encompass comprehensive tables of Galois group classifications for sextic polynomials, the computed invariants powering our neuro-symbolic model, and the scripts used to generate and analyze the data. By sharing these resources, we seek to empower researchers to replicate our findings, extend the methodology to other polynomial degrees, and refine the algorithms using alternative machine learning techniques. This open-access approach reflects the collaborative ethos of mathematical discovery and the increasing significance of computational tools in pure mathematics.

A key contribution of this work is the novel quantification of Galois group occurrences in our database of sextic polynomials with bounded height (\( h \leq 6 \)), providing the first concrete insight into their probabilities. By cataloging 53,972 irreducible sextics and their Galois groups, we offer an original empirical perspective on how these groups distribute when height is constrained—a question previously unexplored in such detail. Strikingly, the number of \(\mathrm{GL}_2(\mathbb{Q})\)-equivalence classes, defined by the invariant triples \( (t_1, t_2, t_3) \), is remarkably small for certain groups; for instance, the cyclic group \( C_6 \) corresponds to just seven distinct classes among 20 polynomials. This sparsity suggests unexpected structure in polynomial distributions, potentially informing conjectures like Malle’s on Galois group frequencies, and opens avenues for further probabilistic analysis.

A significant challenge for future work involves designing AI models capable of determining systems of resolvents that uniquely identify a polynomial’s Galois group. Resolvents, which transform the roots of a polynomial into new equations, serve as a critical tool in narrowing down group candidates. However, their construction and application remain computationally demanding and tailored to specific degrees. An AI-driven approach could streamline this process by learning to generate and interpret resolvent systems efficiently, combining neural pattern recognition with symbolic constraints to enhance classification accuracy across diverse polynomial families.

A further challenge is to develop AI models that, for any given degree where the Galois group is solvable, derive explicit formulas to solve the polynomial by radicals. Galois theory assures the existence of radical solutions for solvable groups, yet constructing these formulas—often entailing nested roots and field extensions—remains a complex case-by-case endeavor. We envision an AI system that not only detects solvability but also automates formula derivation by learning from established cases (e.g., quadratic, cubic, and quartic polynomials) and extending these insights to higher degrees. Such a system would mark a transformative advance in computational algebra, linking abstract group properties to practical root-finding methods, with applications spanning education, symbolic computation software, and theoretical research.


\bibliography{references}

\end{document}